\documentclass{article}
\usepackage{spconf,amsmath,graphicx}
\usepackage{url}
\usepackage{cleveref}
\crefname{figure}{Fig.}{Figs.}   
\Crefname{figure}{Fig.}{Figs.}   
\crefname{table}{Tab.}{Tabs.}   
\Crefname{table}{Tab.}{Tabs.}   
\usepackage{stfloats}
\usepackage{amsfonts}
\usepackage{graphicx}
\usepackage[table]{xcolor}
\usepackage{booktabs}

\usepackage{float}
\usepackage{lipsum}
\usepackage{graphicx}
\usepackage{stfloats}
\usepackage{tabularx}
\usepackage{xstring}
\usepackage{multirow}
\usepackage{xspace}
\usepackage{pgf}
\usepackage{url}
\usepackage{subcaption}
\usepackage[hang,flushmargin]{footmisc}
\usepackage{makecell}
\usepackage{algorithm} 
\usepackage{algpseudocode}
\usepackage[super]{nth}
\usepackage{mathrsfs}
\usepackage{cancel}
\usepackage{times}
\usepackage{microtype}
\usepackage{epsfig}
\usepackage{caption}
\usepackage{float}
\usepackage{placeins}
\usepackage{color, colortbl}
\usepackage{stfloats}
\usepackage{enumitem}
\usepackage{comment}
\usepackage{amssymb}

\usepackage{enumitem}
\setlist{nosep, leftmargin=14pt}

\usepackage{mwe} 

\title{LmPT: Conditional Point Transformer for Anatomical Landmark Detection on 3D Point Clouds}
\name{%
\parbox{\linewidth}{\centering Matteo Bastico\textsuperscript{\textnormal{1\normalsize*}}, Pierre Onghena\textsuperscript{\textnormal{2\normalsize*}}, David Ryckelynck\textsuperscript{\textnormal{3}}, Beatriz Marcotegui\textsuperscript{\textnormal{2}}\\[0.2ex]
\parbox{\linewidth}{\centering Santiago Velasco-Forero\textsuperscript{\textnormal{2}}, Laurent Corté\textsuperscript{\textnormal{1}}, Caroline Robine--Decourcelle\textsuperscript{\textnormal{4}}, Etienne Decencière\textsuperscript{\textnormal{2}} \thanks{\textsuperscript{\normalsize*}Equal contribution}%
}}}
\address{\textsuperscript{1}Mines Paris - PSL University, Centre for material sciences (MAT), 91003 Evry, France \\ \textsuperscript{2}Mines Paris - PSL University, Centre for mathematical morphology (CMM), 77300 Fontainebleau, France \\ \textsuperscript{3}Mines Paris - PSL University, Centre for material forming (CEMEF), 06904 Sophia Antipolis, France \\ \textsuperscript{4}The National Veterinary School of Alfort (EnvA), 94700 Maisons-Alfort, France}
\begin{document}
\pagestyle{plain}
%
\maketitle
\begin{abstract}
Accurate identification of anatomical landmarks is crucial for various medical applications.
Traditional manual landmarking is time-consuming and prone to inter-observer variability, while rule-based methods are often tailored to specific geometries or limited sets of landmarks.
In recent years, anatomical surfaces have been effectively represented as point clouds, which are lightweight structures composed of spatial coordinates.
Following this strategy and to overcome the limitations of existing landmarking techniques, we propose Landmark Point Transformer (LmPT), a method for automatic anatomical landmark detection on point clouds that can leverage homologous bones from different species for translational research.
The LmPT model incorporates a conditioning mechanism that enables adaptability to different input types to conduct cross-species learning.
We focus the evaluation of our approach on femoral landmarking using both human and newly annotated dog femurs, demonstrating its generalization and effectiveness across species.
The code and dog femur dataset will be publicly available at: \url{https://github.com/Pierreoo/LandmarkPointTransformer}.

\end{abstract}

\section{Introduction}\label{sec:introduction}
\begin{figure}[t]
    \centering
    \includegraphics[width=0.335\textwidth]{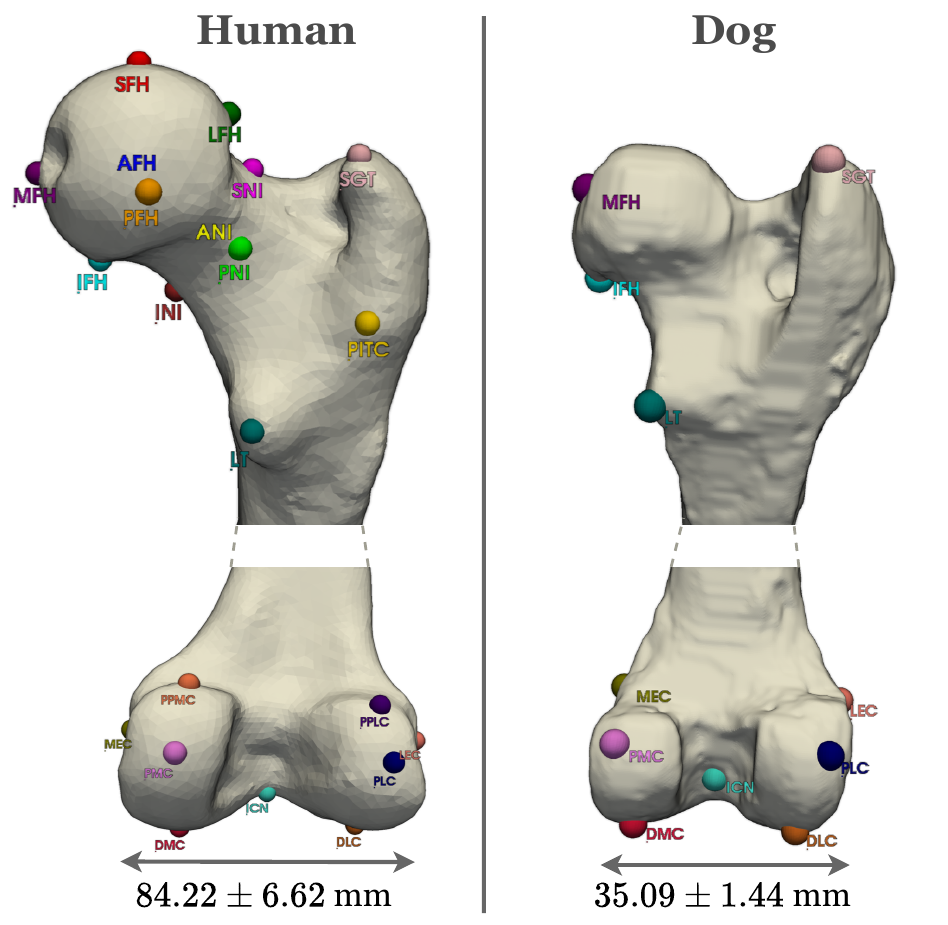}
    \caption{Ground truth landmarks for human and dog femurs.}
    \label{fig:landmarks_gt}
\end{figure}
Anatomical landmark detection is essential in the medical domain for applications like surgical planning and kinematic modeling.
In orthopedics, the accurate detection of landmarks is particularly crucial to assess skeletal deformities and to model joint kinematics. 
Among these, the femoral bone is a central structure due to its biomechanical importance and involvement in several traumatic conditions.
The International Society of Biomechanics (ISB) proposed a standardized femoral coordinate system defined by anatomical landmarks~\cite{wu_isb_2002}.
Alternative landmark-based definitions, such as the Table Top Plane method~\cite{uemura_effect_2019}, have also been introduced.
At the same time, translational research often relies on animal models to investigate disease mechanisms and evaluate new surgical techniques, introducing additional challenges related to cross-species anatomical variability~\cite{OLAH2021151680}.
Reliable identification of key femoral landmarks, including the femoral head extremities, medial and lateral epicondyles, trochanters, and intercondylar notch, is essential (see \cref{fig:landmarks_gt}).
The identification of these landmarks enables a precise definition of the femoral mechanical and anatomical axes for accurate gait analysis, preoperative planning, and 3D modeling~\cite{uemura_effect_2019}.
Traditionally, femoral landmarks have been identified manually by expert radiologists or orthopedic surgeons on medical images from CT or MRI scans, or directly on the 3D models. 
While often considered the standard, manual landmarking is time-consuming, sample-dependent, and shows significant inter‑observer variability~\cite{fischer_robust_2020}.
In contrast, automated methods offer greater scalability, consistency, and efficiency.
Recently, a variety of computational methods have been developed that range from geometric heuristics to statistical shape models and, more recently, data-driven deep learning approaches~\cite{9320194}.
After a review of literature on anatomical landmark identification and keypoint detection in general, we propose a conditional model based on the Point Transformer (PT) architecture~\cite{zhao_point_2021,wu_point_2022,wu_point_2024} to learn landmark locations from different input types, e.g. homologous bones across species.
The contribution of this paper is threefold:
\begin{itemize}
    \item We propose LmPT, a landmark detection method for 3D point clouds that leverages a transformer architecture with conditioning mechanism to adapt to different input types.
    \item We integrate cross-species learning and introduce a new annotated dataset of dog femoral bones to enable translational research settings.
    \item We validate our approach on both human and dog femurs to demonstrate its ability to generalize across homologous femoral structures.
\end{itemize}


\setcounter{figure}{2} 
\begin{figure*}[b]
    \centering
    \includegraphics[width=0.83\textwidth]{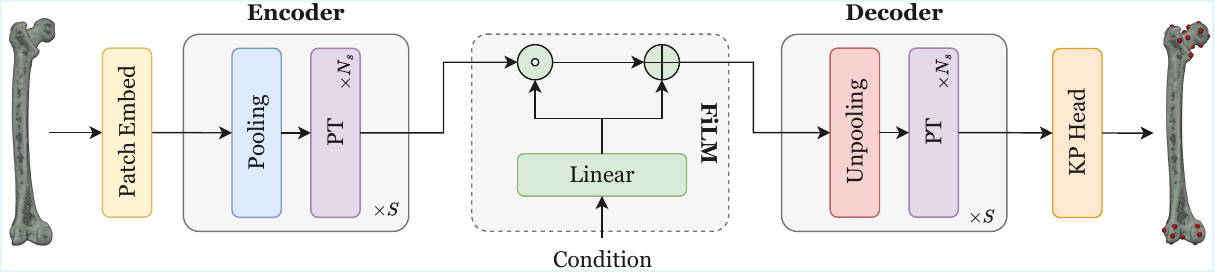}
    \caption{Overview of LmPT, outlining a PT encoder-decoder structure with a FiLM modulation to condition the model by input.}
    \label{fig:overview}
\end{figure*}
\section{Related Work}\label{sec:related}

Identification of medical landmarks can be performed on 2D/3D images or directly on 3D meshes and point clouds, using either rule-based or learning-based approaches.

\noindent\textbf{Rule-based} \quad
Initial attempts on medical images relied on rule-based systems through the use of geometric properties like ridges, corners or saddles. 
More recently, \textit{Fischer et al.}~\cite{fischer_robust_2020} proposed an atlas- and a priori knowledge-based approach, named the A\&A method that is divided in two stages to processes femoral surface models.
Firstly, a single atlas-based registration maps landmarks and areas from a template surface to the subject. 
In the second stage, landmarks, axes and planes are used to construct the femoral bone coordinate systems that are refined with a priori knowledge. 
\textit{Fung et al.}~\cite{10.1007/978-3-031-30111-7_53} predicted anatomical landmarks by non-rigidly deforming an annotated source model to a target point cloud.
These approaches proved effective in structured anatomical regions with few annotated landmarks but suffer when more generalization is needed. 
In fact, they are often limited to a subset of landmarks that are tailored to specific parts or species, and do not generalize well across different geometries.

\noindent\textbf{Learning-based} \quad
Several deep learning approaches, such as classifiers, heatmap regression networks, U-Net variants, multi-view networks, and transformer-based networks have been proposed, but limited work has been applied to 3D point clouds.
Among them, \textit{Sullivan et al.}~\cite{9320194} leverage a hierarchical network to segment input point clouds into background and landmark regions. Offset vectors are then calculated within the landmark regions to refine the predicted landmark locations.
More generally, outside the medical domain, \textit{You et al.}~\cite{9157559} laid the foundation with KeypointNet as a 3D keypoint dataset of generic objects. They evaluate various deep learning backbones such as DGCNN~\cite{10.1145/3326362} for keypoint prediction in both saliency and correspondence tasks.
However, their approach is restricted to single-category training, making it unsuitable for cross-species or cross-category generalization.

\section{Method}\label{sec:method}
Point Transformer (PT) architectures~\cite{zhao_point_2021, wu_point_2022, wu_point_2024} are tailored for point cloud perception. 
PTv2~\cite{wu_point_2022} adopts an encoder–decoder structure with pooling and unpooling between transformer layers to reduce and restore point cloud resolution.
The self-attention mechanism enables the model to focus on relevant neighboring points to capture complex relationships. 
Enlarging (or reducing) the $k$-NN attention windows across $S$ encoder (decoder) layers facilitates multi-scale feature learning, each with $N_s$ transformer layers, as illustrated in~\cref{fig:attention_windows}. 
In the latest version, PTv3~\cite{wu_point_2024} replaces the neighborhood operations with an efficient point cloud serialization scheme based on space-filling curve patterns.
\setcounter{figure}{1}
\begin{figure}[ht!]
    \centering
    \includegraphics[width=0.49\textwidth]{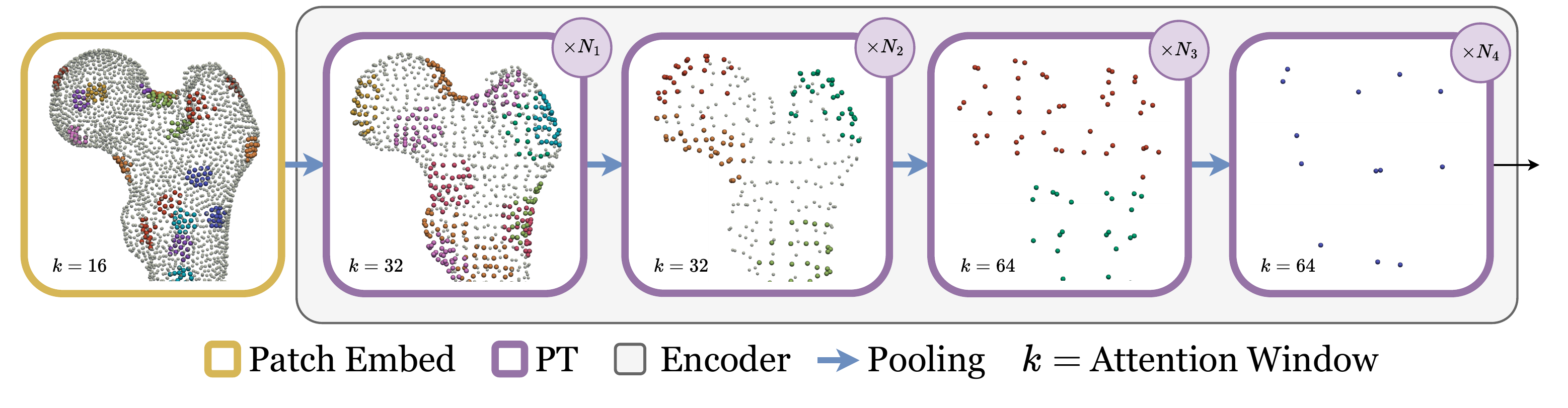}
    \caption{Overview of femur processing with the PTv2 encoder.}
    \label{fig:attention_windows}
\end{figure}

\noindent\textbf{LmPT} \quad
The proposed LmPT framework is designed to detect landmarks with the leverage of PT processing capabilities. 
The general architecture in~\cref{fig:overview} consists of a transformer-based encoder-decoder structure that is effective for tasks requiring fine-grained precision such as landmark detection. 
Finally, the keypoint prediction head (KP Head) produces logits that identify the predicted landmark for each keypoint class based on the encoded features.
Differently from the original PT architecture, a modulation of bottleneck features is introduced to condition the model on the input type. 
To that purpose, the model employs a Feature-wise Linear Modulation (FiLM)~\cite{10.5555/3504035.3504518}, which learns to adaptively influence the output of a neural network by applying a transformation to the network’s intermediate features. 
More formally, a linear layer is used to learn scale and shift parameters for feature-wise affine transformation. 
In the LmPT workflow, modulation is applied only to the bottleneck features of the encoder to reduce model overhead. 
As demonstrated in the experiments section, the conditioning mechanism allows the model to generalize the prediction of landmarks across different species.
Therefore, under conditions of limited data or in translational research settings, we aim to incorporate homologous bones to improve the detection performance.

\section{Experiments}\label{sec:experiment}

\subsection{Experimental Setup}
\noindent\textbf{Datasets} \quad
The methods are evaluated and compared on two annotated datasets of landmarks: the human public femoral landmarks dataset from \textit{Fischer et al.}~\cite{fischer_robust_2020} and our internally collected dog femoral landmarks dataset.
An annotated femur example is visualized in~\cref{fig:landmarks_gt} for each dataset.
The dog landmarks closely correspond to their human counterparts, with some differences due to anatomical variations between species.
The human dataset includes 20 models of human femurs from different patients, ten left and ten right, with 22 anatomical landmarks annotated per model.
Five experts independently positioned the landmarks, each processing all models and repeating the estimation four times, resulting in a total of 20 annotation rounds per femur.
To define a single ground-truth point for our experiments, we computed the medoid of the 20 annotations for each landmark.
The twenty femur models are divided into 16 train and four test samples, with the training set used to fit the proposed models and the test set to evaluate their performance.
Our new dog dataset includes 14 models of dog femurs from different breeds and sizes, seven left and seven right, each segmented from CT scans and annotated by a veterinary expert with 11 landmarks whose subset is also present in the human dataset.
The 14 femur models are split into ten train and four test samples.
Consequently, the combined human and dog dataset provides 26 train and eight test samples to optimize our cross-species model and evaluate its ability to learn from homologous bones across species.

\vspace{0.5em}
\noindent\textbf{Metrics} \quad
The benchmark follows standard metric evaluation practices employed in other detection methods~\cite{fischer_robust_2020, 9157559} to ensure a fair comparison. 
Localization accuracy is assessed using the mean absolute error (MAE), computed as the Euclidean distance between the predicted and ground truth landmarks. 
Additionally, the percentage of correct keypoints (PCK) measures whether predicted landmarks fall within a specified Euclidean distance threshold from the ground truth, evaluated here across ten thresholds ranging from 1mm to 8mm.

\newpage
\noindent\textbf{Implementation Details} \quad
To address data scarcity, point clouds are normalized, uniformly sampled to 8192 points, and augmented through random rotation, scaling, and coronal-plane flipping to enhance cross-side generalization on symmetric landmarks. 
All models are trained under identical conditions for 500 epochs with a batch size of 4, using a channel-wise cross-entropy loss that ignores unlabeled keypoints.
The network employs a AdamW optimizer~\cite{loshchilov_decoupled_2019}, learning rate of $3 \times 10^{-4}$, and coupled with a one-cycle learning rate scheduler.


\subsection{Landmark Detection}
As a proof of concept, we first assess our method on the human femoral dataset to compare predicted landmark accuracy against clinician annotations. 
We then extend the analysis to our dog dataset that includes a single manual annotation of only a subset of human landmarks. 
Finally, we evaluate the cross-species generalization of the LmPT model trained simultaneously on both human and dog femurs.

\setcounter{figure}{3} 
\begin{figure}[ht!]
    \centering
    \includegraphics[width=0.5\textwidth]{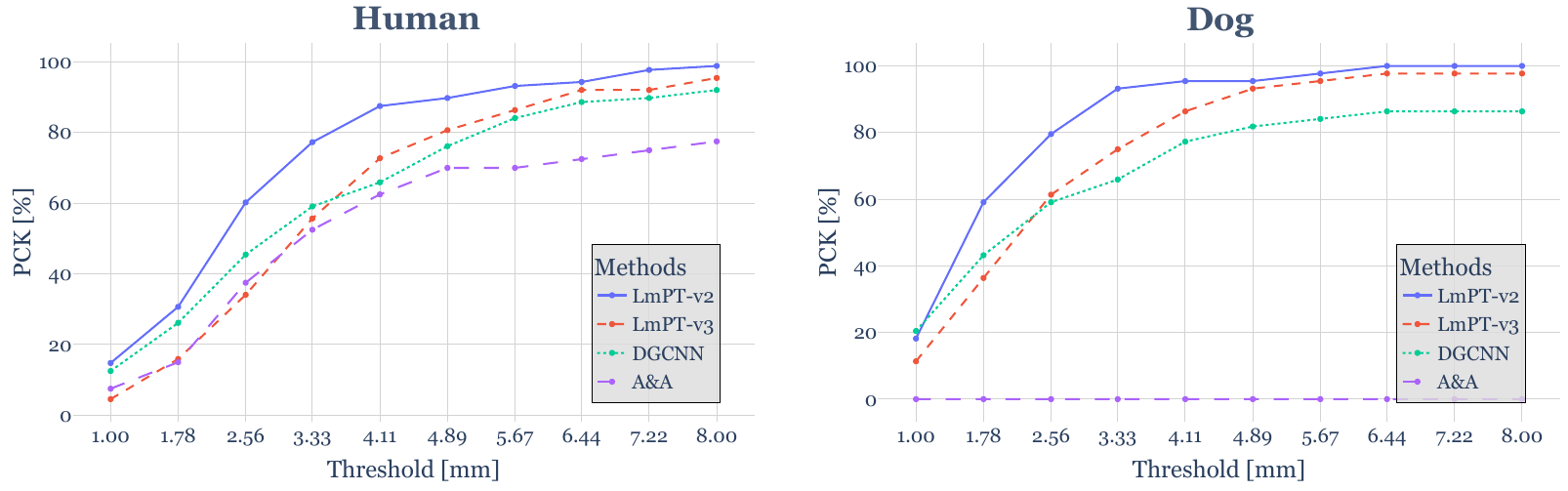}
    \caption{Human (left) and dog (right) femur dataset, PCK (\%).}
    \label{fig:pck}
\end{figure}

\begin{figure}[ht!]
    \centering
    \includegraphics[width=0.48\textwidth]{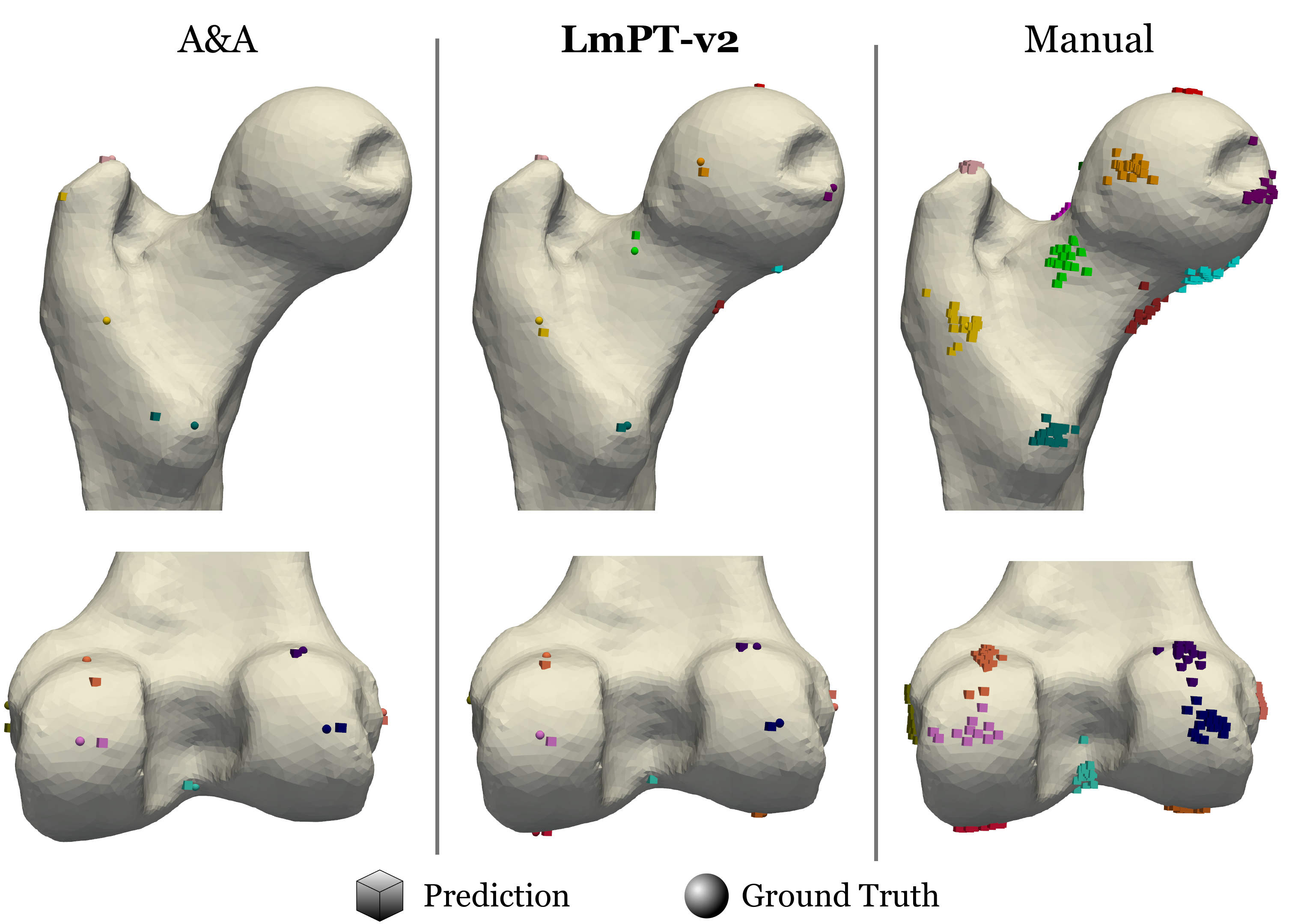}
    \caption{Method comparison of human landmark predictions.}
    \label{fig:visualization_human}
\end{figure}

\noindent\textbf{Human} \quad
The detection results on human femurs are presented in~\cref{tab:human_results}, and include only the landmarks that are analyzed by the A\&A~\cite{fischer_robust_2020} method.
For reference, the average and maximum errors of the manual annotations relative to their medoid are reported as the average manual error (AME) and maximum manual error (MME), respectively.
The proposed LmPT model with the PTv2 backbone surpasses expert-level accuracy on the majority of individual landmarks to achieve a minimal MAE.
Interestingly, the LmPT model with the PTv3 backbone performs worse than PTv2 because PTv3’s serialized mapping strategy for attention windows, while more computationally efficient, is less precise than PTv2’s $k$-nearest neighbors mechanism.
In the left graph of~\cref{fig:pck}, the PCK metric across thresholds highlights the stronger performance of the LmPT model with PTv2.
A qualitative comparison of landmark predictions between A\&A and LmPT-v2 is shown in~\cref{fig:visualization_human}, with LmPT-v2 more closely aligned to the medoids of manual annotations.

\begin{table}[ht!]
\centering
\resizebox{0.48\textwidth}{!}{
\begin{tabular}{c|cccccccccc|c}
\Xhline{1.5pt}
\textbf{Method} & SGT & LT & PITC & PLC & PMC & PPLC & PPMC & LEC & MEC & ICN & Mean \\
\Xhline{1.5pt}
\textbf{AME} & 4.10 & 2.87 & 4.91 & 2.80 & 2.66 & 3.63 & 3.91 & 2.50 & 5.14 & 3.08 & 3.56 \\
\textbf{MME} & 18.77 & 8.50 & 14.64 & 10.58 & 10.06 & 11.20 & 13.99 & 9.05 & 13.24 & 9.86 & 11.99 \\
\Xhline{0.5pt}
\textbf{A\&A} & 6.94 & 6.32 & 29.32 & \textbf{1.76} & 3.37 & 3.88 & 4.10 & \textbf{1.37} & 5.37 & 2.71 & 6.51 \\
\textbf{DGCNN} & \textbf{1.44} & 2.11 & 3.36 & 10.01 & 2.87 & 8.74 & 2.88 & 2.89 & 13.81 & 1.88 & 5.00 \\
\cellcolor{gray!20}\textbf{LmPT-v3} & \cellcolor{gray!20}3.41 & \cellcolor{gray!20}3.86 & \cellcolor{gray!20}4.39 & \cellcolor{gray!20}2.69 & \cellcolor{gray!20}3.68 & \cellcolor{gray!20}\textbf{3.55} & \cellcolor{gray!20}2.89 & \cellcolor{gray!20}4.86 & \cellcolor{gray!20}5.96 & \cellcolor{gray!20}3.38 & \cellcolor{gray!20}3.87 \\
\cellcolor{gray!20}\textbf{LmPT-v2} & \cellcolor{gray!20}1.58 & \cellcolor{gray!20}\textbf{1.57} & \cellcolor{gray!20}\textbf{2.98} & \cellcolor{gray!20}3.05 & \cellcolor{gray!20}\textbf{2.33} & \cellcolor{gray!20}3.67 & \cellcolor{gray!20}\textbf{2.40} & \cellcolor{gray!20}2.90 & \cellcolor{gray!20}\textbf{3.31} & \cellcolor{gray!20}\textbf{1.62} & \cellcolor{gray!20}\textbf{2.54} \\
\Xhline{1.5pt}
\end{tabular}}
\caption{Comparison on human femur dataset, MAE (mm).}
\label{tab:human_results}
\end{table}


\noindent\textbf{Dog} \quad
From the results in~\cref{tab:dog_results} and the performance curves in the right graph of~\cref{fig:pck}, similar findings can be observed in the experiments conducted solely on the dog femoral dataset.
The proposed LmPT with the PTv2 backbone outperforms the other methods, achieving the lowest MAE and the fastest convergence to perfect PCK.
The A\&A~\cite{fischer_robust_2020} method is not reported, as its initial atlas registration step failed for all test samples, an expected outcome given that the method is designed for human femurs and does not generalize well to other species.

\begin{table}[ht!]
\centering
\resizebox{0.48\textwidth}{!}{
\begin{tabular}{c|ccccccccccc|c}
\Xhline{1.5pt}
\textbf{Method} & MFH & IFH & SGT & LT & PLC & PMC & LEC & MEC & ICN & DLC & DMC & Mean \\
\Xhline{1.5pt}
\textbf{DGCNN} & 3.95 & 2.05 & 2.24 & 1.66 & 8.23 & 7.99 & 17.75 & \textbf{2.97} & 1.73 & 5.90 & 1.27 & 5.07 \\
\cellcolor{gray!20}\textbf{LmPT-v3} & \cellcolor{gray!20}2.85 & \cellcolor{gray!20}3.01 & \cellcolor{gray!20}1.92 & \cellcolor{gray!20}2.51 & \cellcolor{gray!20}\textbf{1.10} & \cellcolor{gray!20}1.96 & \cellcolor{gray!20}2.49 & \cellcolor{gray!20}3.57 & \cellcolor{gray!20}1.67 & \cellcolor{gray!20}6.87 & \cellcolor{gray!20}1.74 & \cellcolor{gray!20}2.70 \\
\cellcolor{gray!20}\textbf{LmPT-v2} & \cellcolor{gray!20}\textbf{2.72} & \cellcolor{gray!20}\textbf{1.23} & \cellcolor{gray!20}\textbf{1.69} & \cellcolor{gray!20}\textbf{1.41} & \cellcolor{gray!20}1.53 & \cellcolor{gray!20}\textbf{1.79} & \cellcolor{gray!20}\textbf{1.43} & \cellcolor{gray!20}3.37 & \cellcolor{gray!20}\textbf{0.98} & \cellcolor{gray!20}\textbf{1.45} & \cellcolor{gray!20}\textbf{1.20} & \cellcolor{gray!20}\textbf{1.71} \\
\Xhline{1.5pt}
\end{tabular}}
\caption{Comparison on dog femur dataset, MAE (mm).}
\label{tab:dog_results}
\end{table}

\begin{figure}[t]
    \centering
    \includegraphics[width=0.4714\textwidth]{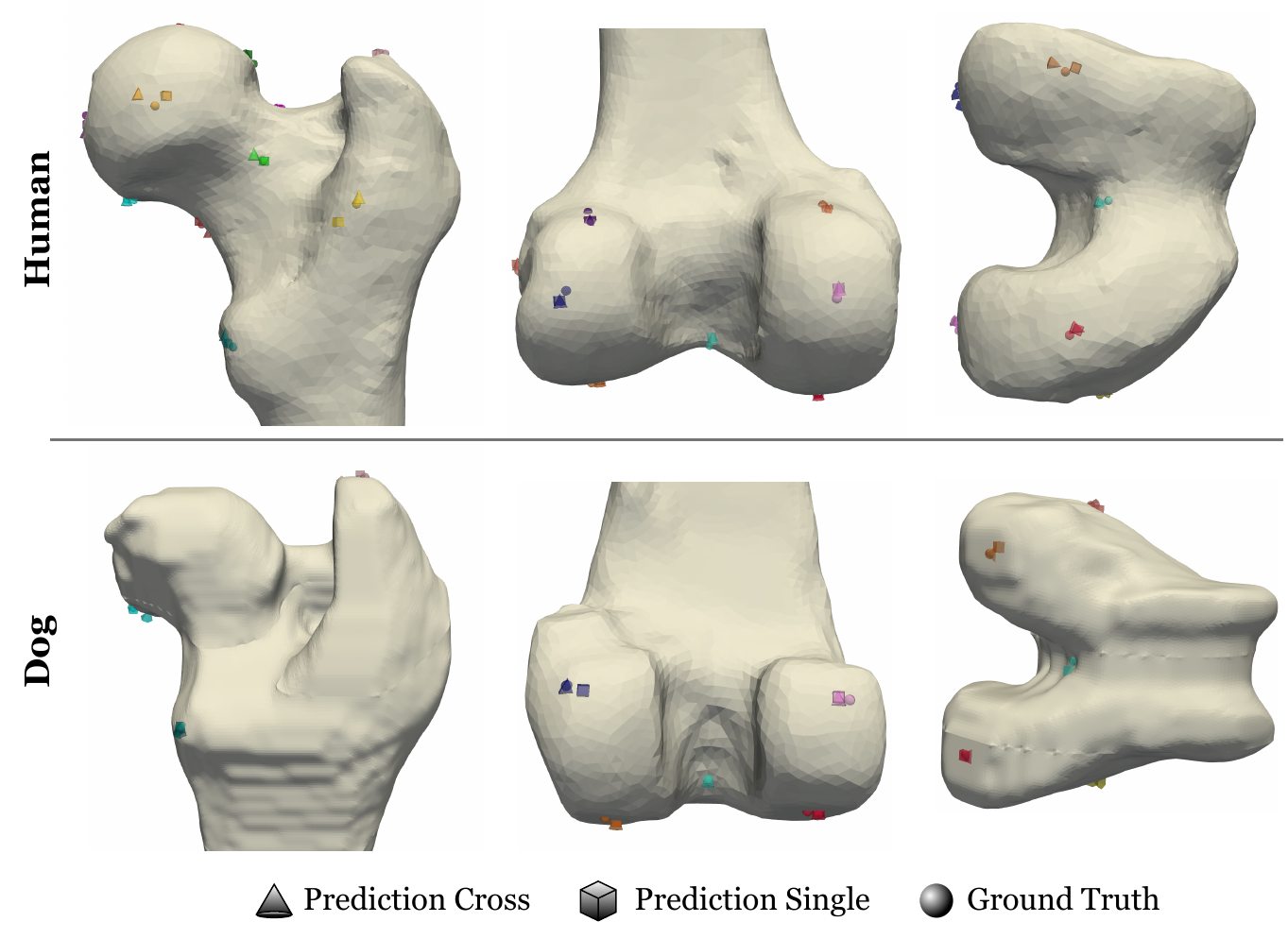}
    \caption{Single- vs cross-species predictions of LmPT-v2.}
    \label{fig:visualization_cross}
\end{figure}

\begin{figure}[ht!]
    \centering
    \includegraphics[width=0.28\textwidth]{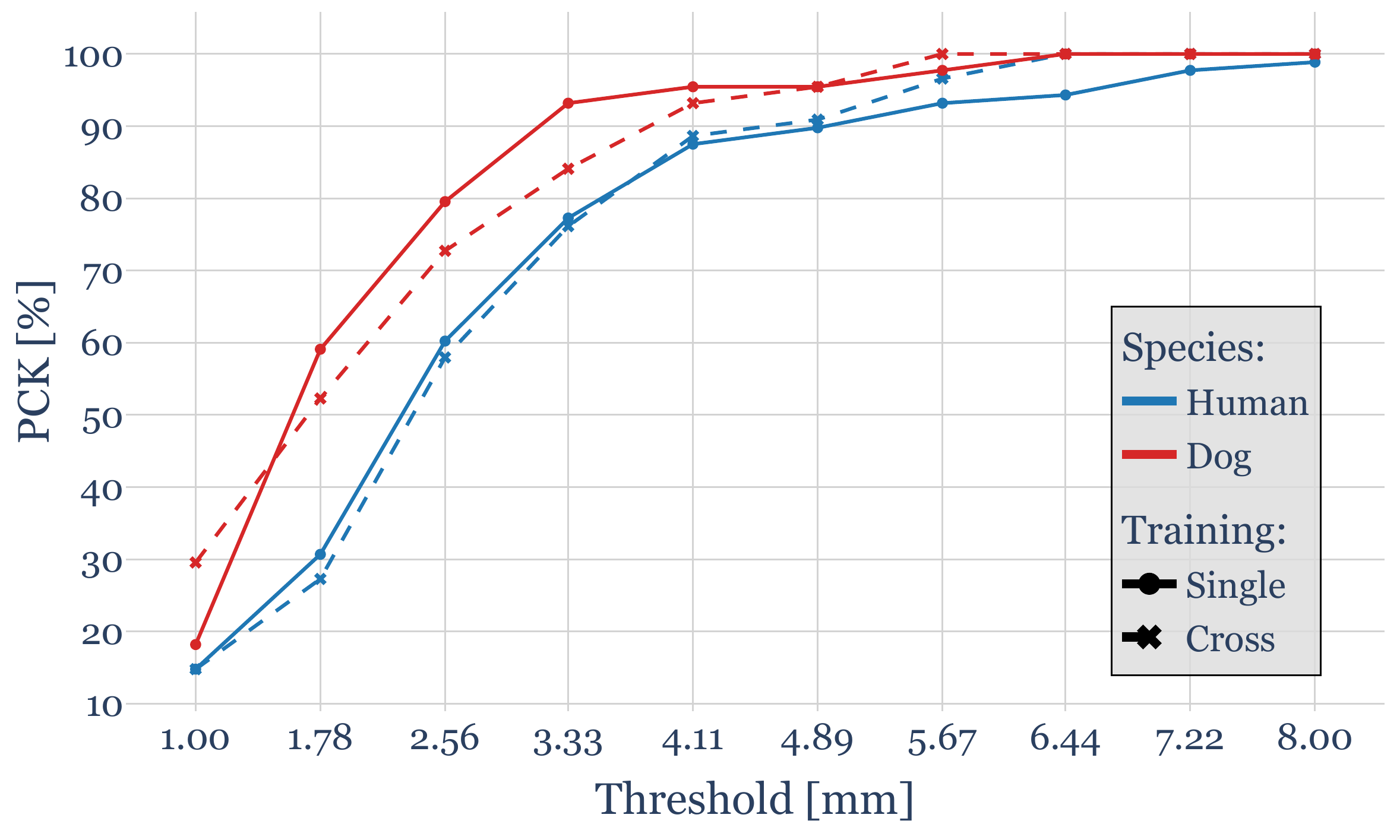}
    
    \vspace{0.5em}
    
    \resizebox{0.48\textwidth}{!}{
    \begin{tabular}{c|c|ccccccccc|c}
    \Xhline{1.5pt}
    \textbf{Species} & \textbf{Cross} & IFH & SGT & LT & PLC & PMC & LEC & MEC & DLC & DMC & Mean \\
    \Xhline{1.5pt}
    \multirow{2}{*}{\textbf{Human}} 
    & \cellcolor{gray!20} & \cellcolor{gray!20}2.37 & \cellcolor{gray!20}\textbf{1.58} & \cellcolor{gray!20}\textbf{1.57} & \cellcolor{gray!20}3.05 & \cellcolor{gray!20}\textbf{2.33} & \cellcolor{gray!20}2.90 & \cellcolor{gray!20}3.31 & \cellcolor{gray!20}\textbf{2.92} & \cellcolor{gray!20}\textbf{1.23} & \cellcolor{gray!20}2.36 \\
    & \cellcolor{gray!20}$\checkmark$ & \cellcolor{gray!20}\textbf{1.50} & \cellcolor{gray!20}2.03 & \cellcolor{gray!20}2.59 & \cellcolor{gray!20}\textbf{2.42} & \cellcolor{gray!20}2.38 & \cellcolor{gray!20}\textbf{1.94} & \cellcolor{gray!20}\textbf{2.98} & \cellcolor{gray!20}2.98 & \cellcolor{gray!20}1.79 & \cellcolor{gray!20}\textbf{2.29} \\
    \Xhline{0.5pt}
    \multirow{2}{*}{\textbf{Dog}}
    & \cellcolor{gray!20} & \cellcolor{gray!20}\textbf{1.23} & \cellcolor{gray!20}\textbf{1.69} & \cellcolor{gray!20}\textbf{1.41} & \cellcolor{gray!20}1.53 & \cellcolor{gray!20}1.79 & \cellcolor{gray!20}1.43 & \cellcolor{gray!20}\textbf{3.37} & \cellcolor{gray!20}\textbf{1.45} & \cellcolor{gray!20}1.20 & \cellcolor{gray!20}\textbf{1.68} \\
    & \cellcolor{gray!20}$\checkmark$ & \cellcolor{gray!20}1.83 & \cellcolor{gray!20}2.41 & \cellcolor{gray!20}1.57 & \cellcolor{gray!20}\textbf{1.50} & \cellcolor{gray!20}\textbf{1.31} & \cellcolor{gray!20}\textbf{0.74} & \cellcolor{gray!20}3.98 & \cellcolor{gray!20}1.52 & \cellcolor{gray!20}\textbf{1.15} & \cellcolor{gray!20}1.78 \\
    \Xhline{1.5pt}
    \end{tabular}}
    
    \caption{Performance of LmPT-v2 in single- and cross-species settings, with PCK (\%) in figure and MAE (mm) in table.}
    \label{fig:cross_species_combined}
\end{figure}

\noindent\textbf{Cross-Species} \quad
The LmPT with the introduced FiLM modulation demonstrates high adaptability in a cross-species setting when trained on both human and dog femurs to achieve results comparable to or surpassing those from single-species training.
Alongside the single-species results of LmPT-v2 in~\cref{tab:human_results,tab:dog_results}, we report the performance of cross-species training in the table of~\cref{fig:cross_species_combined}. 
For the human femurs, the cross-species training decreases the MAE compared to single-species training, indicating that the model benefits from the additional homologous data.
Conversely, for the dog femurs, the cross-species training results in a marginal increase in MAE.
This outcome can be attributed to the difference in the number of landmarks between the two species, with the dog dataset containing only a subset of the landmarks present in the human dataset.
In particular, the human dataset includes additional landmarks that are not directly applicable to the dog anatomy, introducing irrelevant features that confound the model during cross-species training.
However, the dog dataset that contains landmarks shared with the human dataset offers homologous information to progress landmark detection for humans.
The improvement is further supported by the graph of~\cref{fig:cross_species_combined}. For both human and dog categories, single- and cross-species training produce similar curves, but cross-species reaches a perfect score at a lower threshold.
A qualitative comparison of predictions between single- and cross-species is shown in~\cref{fig:visualization_cross}, where the proposed LmPT consistently yields highly accurate landmarks on both species.

\section{Conclusion}\label{sec:conclusion}
In this paper, we addressed the task of precise anatomical landmark detection on point clouds by introducing LmPT, a novel conditional point transformer architecture.
This design enables the adaptation to diverse structures and leverage homologous bones for cross-species translational research.
With the introduction of a newly annotated dataset of dog femurs and extensive experiments, we demonstrate a state-of-the-art in automated femoral landmark detection.
Extending the framework to additional species in future work may further enhance generalization.
Moreover, this work highlights the potential of point cloud learning in the medical domain encouraging its integration into broader automated analysis pipelines.


\clearpage
\section{Compliance with Ethical Standards}\label{sec:compliance}
This research study was conducted retrospectively using human subject data made available in open access by \textit{Fischer et al.}~\cite{fischer_robust_2020}.
Ethical approval was not required as confirmed by the license attached with the open access data.
The dog limbs were explanted from dogs euthanized at the National Veterinary School of Alfort for reasons unrelated to the study and whose owners donated the bodies.
The entire experiment was conducted in accordance with the regulations governing animal experimentation (86/609/CEE, 1986; 2010/63/UE, 2010).

\section{Acknowledgements}\label{sec:acknowledgement}
This work was performed using HPC resources from GENCI–\\IDRIS (Grant 2024-AD011014750R1 and Grant 2024-AD011013902R2).
This project has received funding from the European Union’s Horizon 2020 research and innovation programme under the Marie Skłodowska-Curie grant agreement No945304, AI4theSciences hosted by PSL University.

\bibliographystyle{IEEEbib}
\bibliography{refs}

\end{document}